\documentclass[singlecolumn]{article}
\usepackage{spconf,amsmath,graphicx}

\def\Fig#1{{Fig.\ \ref{fig:#1}}}
\def\Eq#1{{Eq.\ \ref{eq:#1}}}

\title{Fine-grained lesion annotation in CT images with knowledge mined from radiology reports}

\name{Ke Yan$^{\star}$ \qquad Yifan Peng$^{\dagger}$ \qquad Zhiyong Lu$^{\dagger}$ \qquad Ronald M. Summers$^{\star}$}

\address{$^{\star}$ Imaging Biomarkers and Computer-Aided Diagnosis Lab, Radiology and Imaging Sciences, \\National Institutes of Health (NIH) Clinical Center \\
$^{\dagger}$National Center for Biotechnology Information, National Library of Medicine, NIH}
\begin{document}
%
\maketitle
\begin{abstract}
In radiologists' routine work, one major task is to read a medical image, e.g., a CT scan, find significant lesions, and write sentences in the radiology report to describe them. In this paper, we study the lesion description or annotation problem as an important step of computer-aided diagnosis (CAD). Given a lesion image, our aim is to predict multiple relevant labels, such as the lesion's body part, type, and attributes. To address this problem, we define a set of 145 labels based on RadLex to describe a large variety of lesions in the DeepLesion dataset. We directly mine training labels from the lesion's corresponding sentence in the radiology report, which requires minimal manual effort and is easily generalizable to large data and label sets. A multi-label convolutional neural network is then proposed for images with multi-scale structure and a noise-robust loss. Quantitative and qualitative experiments demonstrate the effectiveness of the framework. The average area under ROC curve on 1,872 test lesions is 0.9083.
\end{abstract}
\section{Introduction}
\label{sec:intro}

Lesion detection and classification are important research topics in computer-aided diagnosis (CAD) \cite{Esteva2017nature, Diamant2016liver, Yan2018DeepLesion}. Existing lesion classification studies typically focus on certain body parts (e.g.\ skin \cite{Esteva2017nature} and liver \cite{Diamant2016liver}) and classify the types of lesions (e.g. cyst, metastasis, and hemangioma in liver \cite{Diamant2016liver}). In this paper, we tackle a more general problem to mimic radiologists. When an experienced radiologist reads a medical image such as a computed tomography (CT) scan, he or she can find all kinds of lesions in various body parts, and tell the lesions' detailed body part, type, and attributes. We aim to develop an algorithm to predict these characteristics assuming that the lesion has been detected or manually marked on an image. In brief, we hope the computer will comprehensively understand the lesion and answer the question ``what is it?''. We call it lesion annotation as it is similar to the multi-label image annotation problem in computer vision. We expect it to be an important step towards a fully automated CAD system.

The main challenge of this task is the lack of training labels. Existing lesion classification studies \cite{Esteva2017nature, Diamant2016liver} generally need professionals to label the lesions manually, which is accurate but tedious and not scalable. Some researchers leveraged the rich information contained in radiology reports, but there was no lesion-level correspondence between image and text, thus the extracted labels cannot be accurately mapped to specific lesions \cite{Wang2017ChestXray, Tang2018Xray}. Another stream of study directly generates the report according to the whole image with an attention mechanism to switch among lesions \cite{Wang2018TieNet}. We did not explore this direction because it is difficult to assess the usability of computer-generated reports as their quality is seemingly low for practical use. Instead, if we can predict the lesion-describing keywords accurately, the creation of high-quality (structured) reports would be quite straightforward. 

To find lesion-level labels, we leveraged the recently released DeepLesion dataset \cite{Yan2018DeepLesion}. DeepLesion consists of over 30K lesions on a variety of body parts in CT images. Over 20K of them have corresponding sentences in reports indicated by hyperlinks. Examples of lesion images and sentences can be found in \Fig{res_example}. We collected a fine-grained label list based on the RadLex lexicon \cite{Langlotz2006RadLex}, extracted these labels from the sentences that contain the hyperlinks, and used them as the labels of the lesion image. This process is entirely data-driven and requires minimal manual effort, thus can be easily employed to build large datasets with rich vocabularies.

To improve the coverage of the label list, we added synonyms of each label based on RadLex. The hierarchical relationship between the labels was also annotated and used to expand the label set for each lesion (e.g., before expansion: lung nodule; after expansion: lung nodule, lung, nodule, chest). A multi-label convolutional neural network (CNN) was then adopted to predict all labels for each lesion simultaneously. Since different labels may be best modeled by features at different levels, we modified the structure of the CNN to facilitate multi-scale feature fusion. The loss function was also improved to balance rare labels and mitigate the influence of noisy labels. Experimental results proved our lesion annotator can predict the fine-grained body part, type, and attributes of a variety of lesions with high accuracy.

\section{Label mining from reports}
\label{sec:report_method}

We used the DeepLesion dataset \cite{Yan2018DeepLesion} and its accompanying radiology reports to learn our model. In our hospital, radiologists sometimes mark significant lesions on images and insert hyperlinks, size measurements, or slice numbers (called bookmarks) in reports. Using these bookmarks, we can link the lesion region with the sentence that describes it, and thus obtain lesion-level label annotations. To mine the labels, we first tokenized the sentence that contains the lesion bookmark, and then lemmatized the words in the sentence using NLTK \cite{Bird2016NLTK} to obtain their base forms. RadLex \cite{Langlotz2006RadLex} v3.15 was adopted as our lexicon. We extracted all labels and their synonyms from RadLex. Since most labels in RadLex are nouns, we manually added some adjective synonyms, for example, ``hepatic'' is a synonym of ``liver''. After doing whole-word matching on the sentences and merging the synonyms, we kept the labels with more than 5 occurrences in the test set (1,872 samples), resulting in a list of 145 labels.

\begin{figure}[htb]
	\centering
	\includegraphics[width=8.5cm]{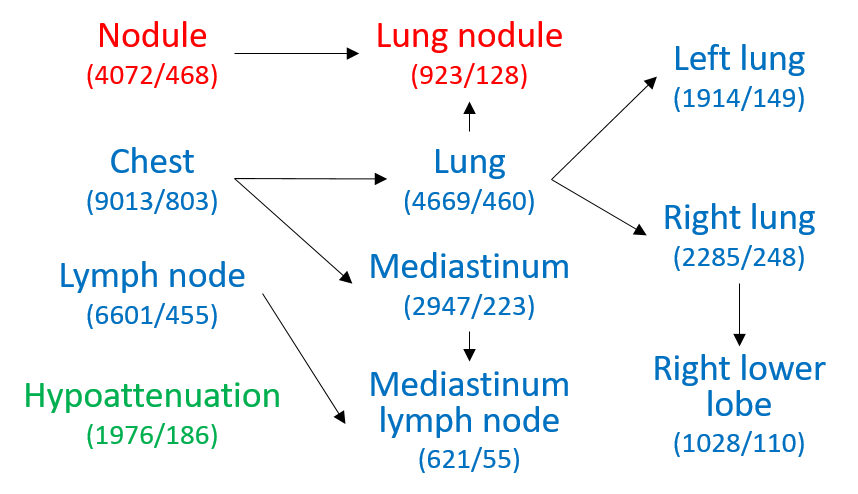}
	\caption{Sample labels with relationships. Arrows point from the parent label to the child. Blue, red, and green indicate body part, type, and attribute, respectively. The numbers below each label are the numbers of training and test samples.}
	\label{fig:term_relation}
\end{figure}

The labels can be categorized into three classes: {\bf 1.\ body parts} (95 labels in total), which include coarse-scale body parts (e.g., chest, abdomen), organs (lung, lymph node), fine-grained organ sub-parts (right lower lobe, retroperitoneum lymph node), and other body regions (porta hepatis, paraspinal); {\bf 2.\ findings / types} (24 labels), which include coarse-scale ones (nodule, mass) and more specific ones (liver mass, ground-glass opacity); {\bf 3.\ attributes} (26 labels), which describe the intensity, shape, size, etc.\ of the lesions (hypoattenuation, spiculated, large). This is a comprehensive set of labels. There are hierarchical relationships between labels, e.g., ``lung nodule'' belongs to ``lung'' and ``nodule''. Therefore, we further extracted the parent-child relationship of the labels from RadLex followed by manual correction. A subgraph of the relationship is shown in \Fig{term_relation}.

The relationship graph was used to expand the label set of a lesion. If a child label is true, all its parent labels should also be true. Then, we can construct a label vector $ \mathbf{y} $ for each lesion. $ y_i=1 $ if and only if label $ i $ is in the expanded label set. It should be noted that in free-text reports, the labels used to describe lesions are often incomplete and in various detail levels. For example, a ``left upper lobe nodule'' may be simply depicted as a ``lung nodule''. It is common to have missing information in the labels. However, we found our algorithm actually worked well with the noisy training labels \cite{Krause2016noisy}, and can be further improved by applying a noise-robust loss function to be introduced in the next section.

\section{Lesion annotation CNN}
\label{sec:image_method}

\begin{figure}[htb]
	\centering
	\includegraphics[width=8.5cm]{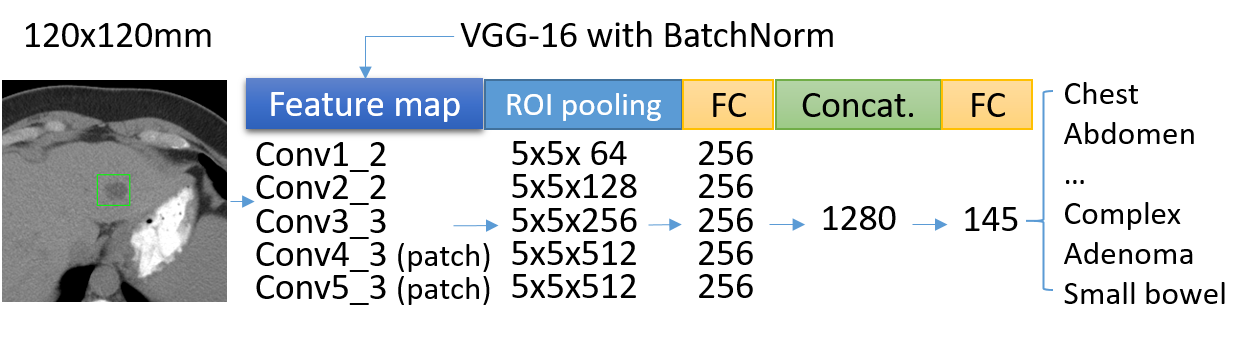}
	\caption{Architecture of the proposed multi-scale multi-label CNN for lesion annotation.}
	\label{fig:network_structure}
\end{figure}

Given a lesion that has been detected or manually marked on an image, our goal is to predict a confidence score for each of the 145 labels. This is a multi-label classification problem. The structure of the proposed CNN is illustrated in \Fig{network_structure}. We first resized the axial CT image to normalize the spacing to 1 mm/pixel. Then, we cropped a fixed-sized 120mm$ \times $120mm patch centered on the lesion. To encode 3D information, we used 3 neighboring slices (interpolated at 2 mm slice intervals) to compose a 3-channel image. The image was sent to a VGG-16 \cite{Simonyan2015Vgg} CNN with batch normalization (VGG-16 showed better accuracy than AlexNet and ResNet50).

One challenge of the task is that different labels may be best modeled by features at different levels. For instance, when predicting intensity and shape attributes, lower-level features are more appropriate. When predicting body parts, the context around the lesion is important, which is contained in higher-level features. Therefore, we combined features from the 5 stages of VGG-16 to get a multi-scale representation. Because of the variable sizes of the lesions, region of interest (ROI) pooling layers were used to pool the feature maps to $ 5 \times  5 \times channel $ separately. For conv1-conv3, the ROI is the bounding box of the lesion in the patch to focus on its details. For conv4 and 5, the ROI is the entire patch to capture the context of the lesion. Each pooled feature map was then passed through a fully-connected layer (FC) and concatenated together. The output of the network is 145 independent scores after another FC layer and a sigmoid function.

Another challenge comes from the extremely imbalanced number of positive labels in different classes. Among the 19,992 training lesions, the most prevailing label (chest) have 9013 positive labels, while the rarest label (internal iliac lymph node) have only 6 positive labels. Generally, coarse-scale body parts and types occur more frequently. If we simply learn a cross-entropy loss for each label, the model will barely learn from rare labels and will always generate low scores for them. A simple solution is to balance the positive and negative cases in each label by assigning different weights, as in Eqs.\ \ref{eq:loss} and \ref{eq:pos_wt}, where $ w_{c,\text{pos}}, w_{c,\text{neg}} $ are the loss weights given to the positive and negative cases of label $ c $, $ N_{c,\text{pos}} $ and $ N_{c,\text{neg}} $ are the numbers of positive and negative training cases of label $ c $, respectively.

Our training labels contain noise since they were extracted from free-text reports. As mentioned in Section \ref{sec:report_method}, there are missing labels in some sentences. In addition, a small number of labels in some sentences are not relevant to the referred lesion. To handle label noise, we replaced the standard cross-entropy loss with the soft bootstrapping loss \cite{Reed2015noisy}. The basic idea is to modify the training label based on the current prediction of the model.

The overall loss function of our model for one sample is shown in \Eq{loss}, where $ K $ is the total number of labels, $ \tilde{y}_{c} $ is the bootstrapped label, $ s_{c} \in [0,1] $ is the score of label $ c $ predicted by the network. In \Eq{bootstrap}, the bootstrapped label is defined as the weighted sum of the original label $ y_c $ and the predicted score $ s_{c} $ \cite{Reed2015noisy}. We empirically set the weight $ \beta=0.9 $ in this paper. If the original label is incorrect and the prediction is close to the actual correct label, the bootstrapped loss will be smaller than the original loss to mitigate the influence of the noisy label. Experimental results show this strategy can improve the prediction accuracy.
\vspace{-1mm}
\begin{equation}\label{eq:loss}
L = \sum_{c=1}^{K} \left(w_{c,\text{pos}} \tilde{y}_{c} \log(s_{c}) + w_{c,\text{neg}}(1-\tilde{y}_{c}) \log(1-s_{c}) \right),
\end{equation}
\begin{equation}\label{eq:pos_wt}
w_{c,\text{pos}} = \frac{N_{c,\text{pos}}+N_{c,\text{neg}}}{2N_{c,\text{pos}}}, w_{c,\text{neg}} = \frac{N_{c,\text{pos}}+N_{c,\text{neg}}}{2N_{c,\text{neg}}},
\end{equation}
\begin{equation}\label{eq:bootstrap}
\tilde{y}_{c} = \beta y_c + (1-\beta)s_{c}.
\end{equation}

\section{Experiments}
\label{sec:exp}

We extracted 19,992 lesion-sentence pairs from DeepLesion for training, and another 1,872 pairs for testing. There was no patient-level overlap between the two sets. The test sentences were manually checked to remove labels that were irrelevant to the corresponding lesions. However, there may still be missing labels, which will require significant amount of manual effort to complete. In this regard, we argue that results on this test set is a nonperfect but reasonable surrogate of the actual accuracy.

The proposed CNN was randomly initialized. It was trained using stochastic gradient descent for 15 epochs with an initial learning rate of 0.01, which was reduced to 0.001 at epoch 12. The batch size was 128. We computed the AUC, i.e.\ the area under the receiver operating characteristic (ROC) curve \cite{Wang2017ChestXray}, for each label on the test set. Table \ref{tbl:acc} displays the results of different methods. In the first row, we evaluated a baseline method by directly attaching a global average pooling layer and an FC layer after VGG-16 with BatchNorm to predict the 145 labels. Then, we added the loss weight in \Eq{pos_wt}. As shown in the second row, the AUC increased significantly. The improvement is larger on rarer labels. We further applied the multi-scale structure in \Fig{network_structure}, and show the results in the third row of Table \ref{tbl:acc}. Prediction accuracies of lesion types and attributes were improved because they rely more on fine details of the lesion, which is contained in mid- and low-level features. Finally, we further implemented the noise-robust bootstrapping loss in \Eq{bootstrap}. The accuracy of lesion attributes was improved the most, possibly because there are many missing attributes in the reports as radiologists typically do not describe every attribute of a lesion in the report. Besides, some attributes may be subjective. Although missing labels are also a problem for fine-grained organ sub-parts, they did not affect the accuracy. It is because organ subparts have relatively stable appearance and are easier to learn. The final overall AUC is $0.9083 \pm 0.0951$.

\begin{table}
	\centering
	\setlength{\tabcolsep}{3pt}
	\renewcommand{\arraystretch}{1.2}
	\begin{tabular}{lllll}
		\hline 
		Method & Overall & Body part & Type & Attribute \\
		\hline
		VGG-16	& 0.8121 & 0.8600 & 0.7618 & 0.6836\\
		 + loss weight & 0.8952 & \bf 0.9443 & 0.8344 & 0.7719 \\
		 + multi-scale	& 0.9045 & \bf 0.9443 & 0.8600 & 0.8000 \\
		 + bootstrap loss	& \bf 0.9083 & 0.9431 & \bf 0.8659 & \bf 0.8203 \\
		\hline
	\end{tabular}
	\caption{AUC averaged on different label subsets.}
	\label{tbl:acc}
\end{table}

To elaborate the results in Table \ref{tbl:acc}, we exhibit the AUCs of some specific labels in Table \ref{tbl:acc_per_term}. Lymph nodes exist all over the body and have variable contextual appearance. Hence, its AUC is lower than its child label, retroperitoneum lymph node, whose appearance is relatively stable. Body regions such as porta hepatis and paraspinal also have stable appearance and high AUC. Similarly, coarse-scale lesion types such as mass and nodule have variable appearance and are slightly subjective, while finer-scale ones like liver mass and ground-glass opacity are more stable, which explains their difference in AUC. Some attributes are subjective or subtle, e.g.\ ``large'' and ``hypoattenuation'', thus attributes have lower average AUC than types and body parts.

\begin{table}
	\centering
	\setlength{\tabcolsep}{3pt}
	\renewcommand{\arraystretch}{1.2}
	\begin{tabular}{p{2.5cm}rr|llr}
		\hline 
		Label & AUC &&& Label & AUC \\
		\hline
		Chest & 0.9426 &&& Nodule & 0.8885 \\
		Abdomen & 0.9414 &&& Mass & 0.8370 \\
		Lung & 0.9750 &&& Liver mass & 0.9690 \\
		Lymph node & 0.9163 &&& Ground-glass opacity & 0.9770 \\
		Right lower lobe & 0.9646 &&& Hypoattenuation & 0.8809 \\
		Retroperitoneum lymph node & 0.9665 &&& Spiculated & 0.8236 \\
		Porta hepatis & 0.9772 &&& Calcified & 0.7213 \\
		Paraspinal & 0.9947 &&& Large & 0.7189 \\
		\hline
	\end{tabular}
	\caption{AUC of typical labels.}
	\label{tbl:acc_per_term}
\end{table}

\begin{figure}[]
	\centering
	\includegraphics[width=8.5cm,trim=90 245 130 50,clip]{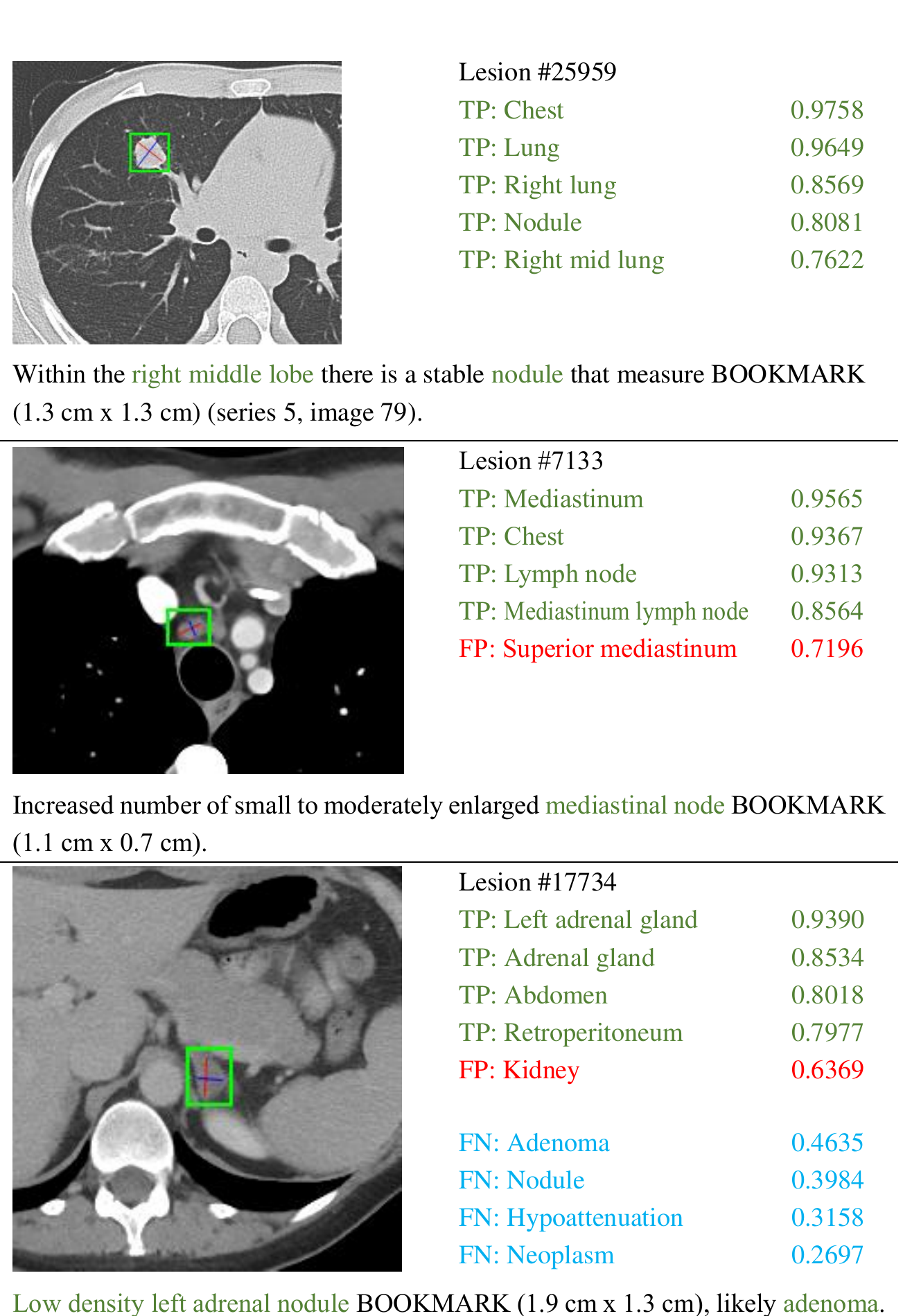}
	\caption{Examples of the lesion image, report sentence, and our predicted labels with scores (based on the image alone) on the test set. The bounding-box and diameters in the image mark the lesion. ``BOOKMARK'' in the report is the hyperlink we used to locate the sentence. We treat the top-5 in the 145 prediction scores of a lesion as its predicted labels. The true labels in top-5 are true positives (TP, in green), false labels in top-5 are false positives (FP, in red), and the true labels not in top-5 are false negatives (FN, in blue).}
	\label{fig:res_example}
\end{figure}

Qualitative results are shown in \Fig{res_example}. In the first example, our algorithm accurately predicted the fine-grained body part and type of a lung nodule. In the second one, a mediastinum lymph node was correctly predicted. The label ``superior mediastinum'' was regarded as false positive, but it is actually a missing true label. In the third example, we identified the lesion's body part: left adrenal gland. Kidney has relatively high score because it is close to the adrenal gland and sometimes confusing for the model. The false negative labels exist possibly due to their variable (nodule, neoplasm) or subtle (hypoattenuation) appearance.

\section{Conclusion}
\label{sec:concl}

In this paper, we mined fine-grained lesion-level description from free-text reports, and then linked them with lesion images to build a comprehensive lesion annotation algorithm. Our future work includes using natural language processing algorithms to reduce label noise in reports \cite{Wang2017ChestXray}, as well as improving prediction accuracy on hard labels and rare labels.

{\bf ACKNOWLEDGMENTS}: This research was supported by the Intramural Research Programs of the National Institutes of Health Clinical Center and National Library of Medicine.

\vspace{-5mm}
\bibliographystyle{IEEEbib}
\bibliography{ISBI19}

\end{document}